\documentclass{article}

 \usepackage[main, final]{neurips_2026}

\usepackage[utf8]{inputenc} 
\usepackage[T1]{fontenc}    
\usepackage{hyperref}       
\usepackage{url}            
\usepackage{booktabs}       
\usepackage{amsfonts}       
\usepackage{nicefrac}       
\usepackage{microtype}      
\usepackage{xcolor}         

\usepackage{enumitem}  

\usepackage{graphicx}
\usepackage{amsmath}

\usepackage[most]{tcolorbox}
\usepackage{xcolor}

\title{Extracting Search Trees from LLM Reasoning Traces Reveals Myopic Planning}

\author{
  Sixing Chen \\
  New York University \\
  \texttt{sixing.chen@nyu.edu}
  \And
  Ji-An Li \\
  New York University \\
  \texttt{jian.li.acad@gmail.com}
  \And
  Saner Cakir \\
  Generality, Inc.\\
  \texttt{saner@generality.inc}
  \And
  Sinan Akcali \\
  Generality, Inc. \\
  \texttt{sinan@generality.inc}
  \And
  Kayla Lee \\
  Generality, Inc. \\
  \texttt{kayla@generality.inc}
  \And
  Marcelo G. Mattar \\
  New York University \\
  \texttt{marcelo.mattar@nyu.edu}
}

\begin{document}

\maketitle

\begin{abstract}
Large language models (LLMs), especially reasoning models, generate extended chain-of-thought (CoT) reasoning that often contains explicit deliberation over possible futures. Yet it remains unclear whether this deliberation constitutes genuine planning, how it is structured, and what aspects of it drive performance. Here we introduce a method for characterizing LLM planning by extracting search trees from reasoning traces in the board game four-in-a-row. By fitting computational models to the extracted search trees, we quantify how plans are structured and how they influence move decisions. We find that LLMs search less deeply than humans, and that their performance is better predicted by search breadth than depth. Most strikingly, although LLMs expand deep nodes in their traces, their move choices are best explained by a myopic model that ignores those nodes entirely. A causal intervention that selectively prunes CoT paragraphs supports that move selection depends predominantly on shallow rather than deep nodes. These patterns contrast with human planning, where performance is driven primarily by deep search. Together, our findings reveal a key difference between LLM and human planning: human expertise rests on deeper search, whereas LLMs generate deep lookahead but fail to act on it. This dissociation offers concrete targets for aligning LLM and human planning. More broadly, our framework provides a generalizable approach for interpreting the structure of LLM planning across strategic domains.
\end{abstract}

\section{Introduction}

Large language models (LLMs), especially reasoning models, have shown a remarkable capability for extended chain-of-thought (CoT) reasoning, in which models generate long reasoning traces before committing to an answer \citep{wei2022chain}. In reasoning models such as DeepSeek-R1 \citep{guo2025deepseek} and OpenAI o1 \citep{openai2024o1systemcard}, these traces can span thousands of tokens and contain explicit deliberation over hypothetical futures. Such deliberation resembles the mental simulation that underlies human planning \citep{li2025system}, raising the possibility that these models engage in genuine prospective planning.

In both classical artificial intelligence (AI) and cognitive science, planning has long been formalized as tree search, with deep forward search regarded as the key driver of planning capability. In AI, game-playing agents such as AlphaGo achieve superhuman performance by systematically searching far into the future \citep{silver2016mastering, silver2018general, schrittwieser2020mastering}. In cognitive science, tree search has likewise served as the primary computational framework for modeling human planning, with evidence that people mentally simulate sequences of future actions to guide their decisions \citep{mattar2022planning, kuperwajs2026looking, callaway2022rational, jensen2024recurrent, collins2026people}. The depth of this simulation scales with expertise: people encountering a new game for the first time plan with fast and flat simulation \citep{collins2026people}, whereas experts search more deeply \citep{van2023expertise, kuperwajs2026looking}.

Whether LLMs engage in comparable search-based planning, however, remains contested. One view holds that LLMs are fundamentally incapable of planning, since autoregressive generation cannot support the systematic search and backtracking that planning requires \citep{kambhampati2024position}. Consistent with this, several behavioral benchmark studies report that LLMs fail at systematic multi-step planning and that their outputs are better explained by pattern completion than genuine planning \citep{valmeekam2025systematic, zheng2024natural}. The opposing position points to the strong performance of reasoning models on challenging tasks that appear to demand multi-step planning, including competitive programming, mathematical reasoning, and strategic gameplay \citep{openai2024o1systemcard, guo2025deepseek, el2025competitive}. However, both positions rest primarily on behavioral outcomes without examining the structure of the reasoning that produced those outcomes.

Resolving this debate therefore requires asking different questions. First, do LLM reasoning traces exhibit the structural hallmarks of systematic search? This question remains largely unaddressed, in part because reasoning traces are long, verbose, and unorganized, making their underlying structures difficult to extract. Recent work has begun to extract structured graphs from reasoning traces to predict reasoning quality, but has focused on tasks with a single correct answer, such as math, science, and coding \citep{jiang2025makes, mukherjee2025premise, zeng2026rejump}. Planning poses a different computational challenge: rather than arriving at a correct answer, it requires evaluating sequences of future actions and their consequences. Second, if LLMs do engage in search, does that search actually drive their decisions? A reasoning trace that looks like search may or may not be what determines the final choice, a gap invisible to behavioral benchmarks and largely unexplored in the existing literature.

\begin{figure}[!t]
    \centering
    \includegraphics[width=\linewidth]{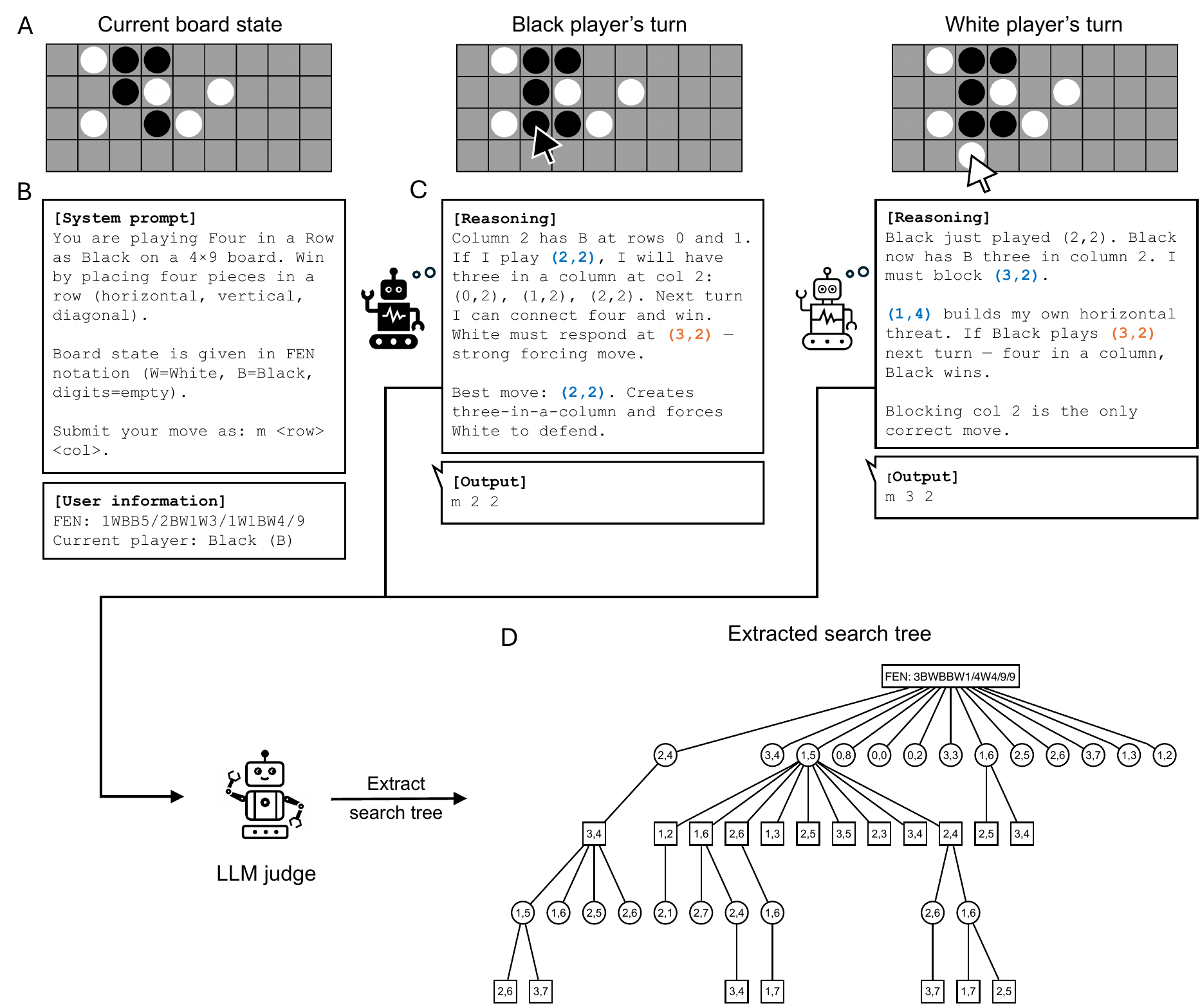}
    \caption{
    \textbf{Game setup and search tree extraction.}
    (A) An example board position in the four-in-a-row game. Two players (black and white) alternate placing pieces on a 4$\times$9 board, and the first player who achieves four-in-a-row wins the game.
    (B) Task prompt. The system prompt describes the rules of four-in-a-row, the board representation in FEN notation, and move submission format. The user message provides the current board state and the active player.
    (C) Reasoning trace and move output. The model generates CoT reasoning traces before committing to a final move in the output. In the example reasoning traces, deliberated moves of the model are highlighted in blue, while deliberated opponent moves are highlighted in orange.
    (D) Search tree extraction. An LLM judge (GPT-5) parses the reasoning trace to extract the search tree of considered moves. In the example search tree, the top square shows the current board state (denoted by the board's FEN notation). Each circle represents a state resulting from a model's own simulated move, and each square represents a state resulting from a simulated opponent's move. Numbers inside each node indicate the coordinate of the corresponding move (zero-indexed). The board, reasoning trace, and search tree in this figure are illustrative only. The search tree does not correspond to the board position shown in (A).
    }
    \label{fig:f1}
\end{figure}

In this work, we address this gap by introducing a method for extracting and quantifying search trees from LLM reasoning traces in a two-player board game. We then fit computational models to characterize how those trees shape move decisions. We focus on the game ``four-in-a-row'' (\autoref{fig:f1}A). This game suits our investigation for several reasons. First, it is a well-defined strategic game, making tree extraction tractable and verifiable. Second, human planning in the game has been characterized by an established computational cognitive model \citep{van2023expertise}, providing a rigorous baseline for direct comparison with human play. Third, popular games like chess or Go are heavily represented in LLM training data, raising the possibility that models rely on memory in these games rather than planning from scratch \citep{pleiss2026trapped, liu2026chessarena}. In contrast, four-in-a-row is less likely to be overrepresented online, making it a cleaner testbed of planning ability.

Analyzing reasoning traces from LLMs playing four-in-a-row, we find that LLMs search more shallowly than humans, and that search depth explains no additional variance in performance once search breadth is controlled for. Crucially, although LLMs do expand deep nodes, their move choices are best explained by a myopic model that ignores those nodes entirely. A causal intervention study in which we selectively prune CoT paragraphs further indicates that move selection depends predominantly on shallow rather than deep search. These patterns contrast with human planning, where expertise rests primarily on deeper search. Together, our findings reveal that LLMs do not act on deep lookahead, and that their planning strategy differs fundamentally from the depth-driven expertise in humans.

\section{Game setup and search tree extraction}

\subsection{Four-in-a-row tournament with LLMs}

We used four-in-a-row to study planning in LLMs. Four-in-a-row is a two-player zero-sum board game (\autoref{fig:f1}A). In this game, two players (white and black) alternate placing pieces on a $4 \times 9$ grid. White moves first. A player wins by placing four of their pieces consecutively along a horizontal, vertical, or diagonal line\footnote{Unlike in the popular game Connect Four, pieces in four-in-a-row are not subject to gravity. A player may place their piece on any empty cell and does not have to drop it to the bottom of a column.}. If the board fills without a winner, the game is a draw.

Each model received a system prompt describing the rules (see \autoref{appendix:methods}.1 for game prompts). Board states were encoded in a FEN-style notation \citep{zhang2025complete, schultz2025mastering}: each row is written as a sequence of piece symbols (\texttt{W} for white, \texttt{B} for black), integers represent runs of consecutive empty cells, and rows are separated by slashes. For example, \texttt{1WBB5/2BW1W3/1W1BW4/9} describes a four-row board in which the first row contains one empty cell, followed by a white piece, two black pieces, and five empty cells (\autoref{fig:f1}A). In each turn, the board state and the player to move were passed as a user message, and the model was asked to respond with a move in the form \texttt{m <row> <col>}, where \texttt{<row>} and \texttt{<col>} are zero-indexed coordinates of the target cell (\autoref{fig:f1}B).

We ran a round-robin tournament among 27 models, with every pair playing four games (alternating which model moves first), yielding a total of 1404 games (see \autoref{appendix:list} for the full list of models). Participating models spanned both proprietary models (e.g., GPT-5, Claude Opus 4.1) and open-weight models (e.g., DeepSeek-R1, Qwen3-235B). Because proprietary models return only summaries of their reasoning traces that omit intermediate reasoning steps, all subsequent analyses were restricted to the 14 models whose complete reasoning traces were accessible. This yielded 9696 reasoning traces across 1092 games.

\subsection{Transcribing reasoning traces into search trees}

Reasoning traces are unstructured natural language, making it difficult to measure planning from them. To address this, we transcribed each trace into a formal search tree using an LLM judge (GPT-5). For each turn, the judge received the model's full response (the concatenation of its reasoning content and output) and was asked to extract every move explicitly deliberated in the reasoning trace into a search tree (\autoref{fig:f1}C--D). In the search tree, coordinates are coded as zero-indexed $(\text{row}, \text{column})$ pairs. Each depth-1 node\footnote{We use \emph{depth} to denote distance from the current board state, which is the root of the search tree. A depth-1 node is the board state after one move by the model, a depth-2 node is the board state after the opponent's reply, and so on. In game terminology, one \emph{ply} is a single move by one player; a $d$-th ply move corresponds to a move that leads to a depth-$d$ state.} represents a candidate first-ply move the model explicitly considered, each depth-2 node represents a reply the model anticipated from the opponent, and so on. The judge produced each search tree as a nested list. For example, the nested list \texttt{[[(2,4), [(1,3)], [(2,2)]], [(0,3)]]} encodes two first-ply moves considered by the model: $(2,4)$ and $(0,3)$. Under $(2,4)$, the model anticipates two opponent replies at $(1,3)$ and $(2,2)$. The other depth-1 node $(0,3)$ is a leaf, meaning the model considers it without further lookahead. Only moves explicitly named in the trace were included; the judge was instructed not to infer or hallucinate moves. This process was applied to all reasoning traces, yielding a structured search tree for each turn.

To develop the extractor, we constructed a human-annotated set of reasoning traces and used it to optimize the extraction prompt before applying the extractor to the full dataset (see \autoref{appendix:methods}.2 for detailed extraction methods). To assess the fidelity of the extractor, we built a separate human-annotated validation set and confirmed that the extracted trees closely match the human-annotated trees (see \autoref{appendix:methods}.3 for the fidelity validation results).

\section{Quantifying search trees extracted from reasoning traces}

\subsection{Search effort predicts win rate}

We first asked whether the amount of search a model performed predicted its game performance. For each model, we computed its average tree size across all turns and its overall win rate in the tournament. Across models, we found a positive relationship between search effort and win rate (\autoref{fig:f2}A), suggesting that models that searched more tended to play better. This relationship also held within model families: within the same model family (e.g., DeepSeek, Qwen, Kimi), models that searched more achieved higher win rates.

A particularly informative case is GPT-OSS-120B, which we ran at two reasoning-effort levels, medium and high. The high-effort condition allocated more tokens to reasoning, resulting in larger search trees and a higher win rate (\autoref{fig:f2}A). Because model architecture, weights, and training were identical across the two conditions, the only difference was the amount of inference-time deliberation. This provides causal evidence that search effort drives the performance gain.

\subsection{LLMs search more shallowly than humans}

\begin{figure}[t]
    \centering
    \includegraphics[width=\linewidth]{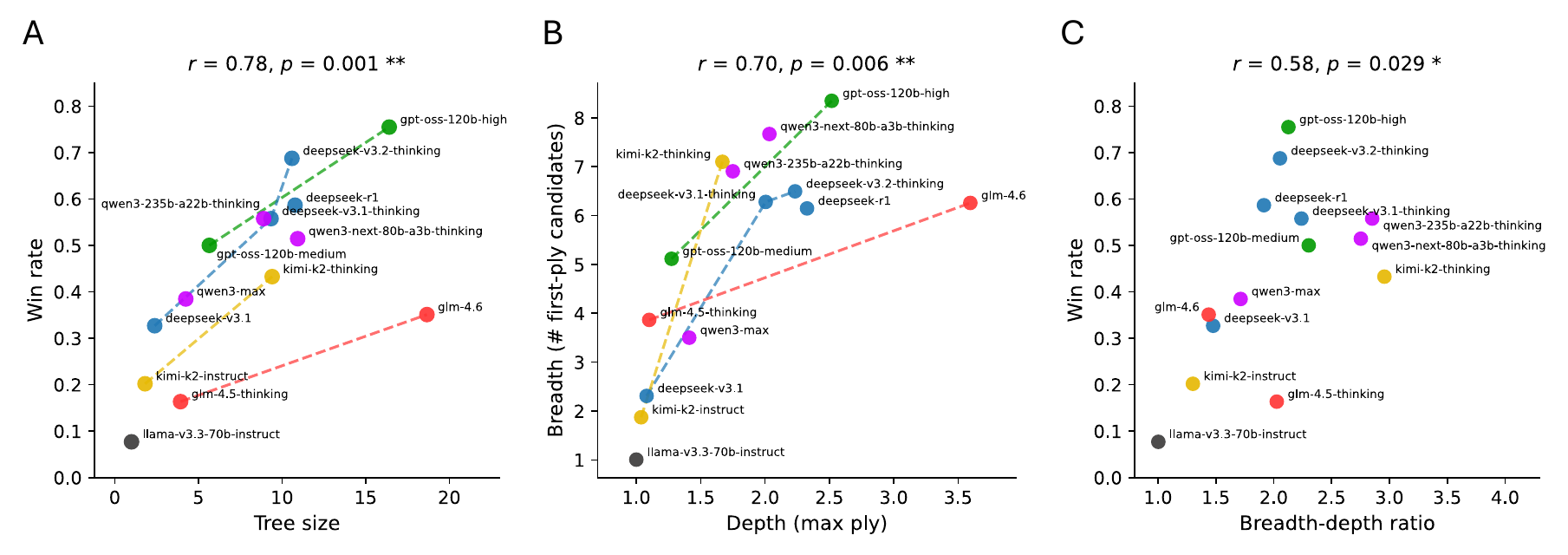}
    \caption{
    \textbf{Planning effort and game performance across LLMs.}
    (A) Win rate as a function of search tree size.
    (B) Search breadth (number of first-ply candidate moves considered) as a function of depth (max ply, i.e., the maximum number of alternating moves simulated ahead) across models.
    (C) Win rate as a function of breadth-depth ratio.
    Dashed lines connect models in a model family. Asterisks denote significance levels ($\text{*} \, p < 0.05$, $\text{**} \, p < 0.01$).
    }
    \label{fig:f2}
\end{figure}

Having established that search effort predicts performance, we next asked which aspect of search accounted for this gain. We considered two dimensions of search effort: depth (the maximum number of steps the model looks ahead) and breadth (the number of candidate moves considered at the first ply). These two dimensions characterize different search strategies: greater depth reflects a tendency to anticipate consequences further into the future (as in depth-first search), whereas greater breadth reflects a tendency to evaluate a wider range of alternatives (as in breadth-first search). Across models, depth and breadth were positively correlated (\autoref{fig:f2}B). The average maximum depth across models ranged between 1.00 and 3.48 plies ($1.78 \pm 0.71$; see \autoref{appendix:results}.1 for depth histograms). This is considerably shallower than the search depth previously inferred from human players in the same game, which falls between 4 and 6 plies and increases with expertise \citep{van2023expertise}.

To quantify the relative contributions of depth and breadth in predicting performance, we fit a linear regression model to predict win rate from both measures. Controlling for breadth, depth was not a significant predictor (depth: $\beta=-0.015, \, p=0.82$; breadth: $\beta=0.075, \, p=0.0025$). Consistent with this, models that explored more first-ply candidates relative to their search depth---reflected in a higher breadth-depth ratio---tended to achieve higher win rates (\autoref{fig:f2}C). Together, these results suggest that LLMs search more shallowly than humans, and the performance advantage of higher-effort models is driven primarily by considering a broader set of candidates rather than by searching deeper.

\section{LLMs do not act on deeper search}

\subsection{Predicting moves from search trees with computational modeling}

To examine how LLMs integrate search tree information into their move decisions, we fit computational cognitive models adapted from the heuristic search model of \citet{van2023expertise}. In their model, candidate moves are evaluated by propagating a heuristic value up a search tree via minimax backup, and the move with the highest backed-up value is selected. Because human search trees are unobservable, \citet{van2023expertise} used the fitted model to infer the latent search trees. In our setting, LLM reasoning traces give us access to the search trees, so we instead predicted move decisions directly from the extracted search trees.

Our cognitive models take an extracted search tree as input and output a probability distribution over candidate first-ply moves. All cognitive models share two core components: a heuristic function that assigns a scalar value to any board state, and a value backup procedure that propagates those values up the tree to score each candidate first-ply move. We held the heuristic function fixed across cognitive models and varied only the value backup rule. This allows us to isolate the backup strategy that best captures how LLMs integrate information across the search tree.

\begin{figure}[!t]
    \centering
    \includegraphics[width=\linewidth]{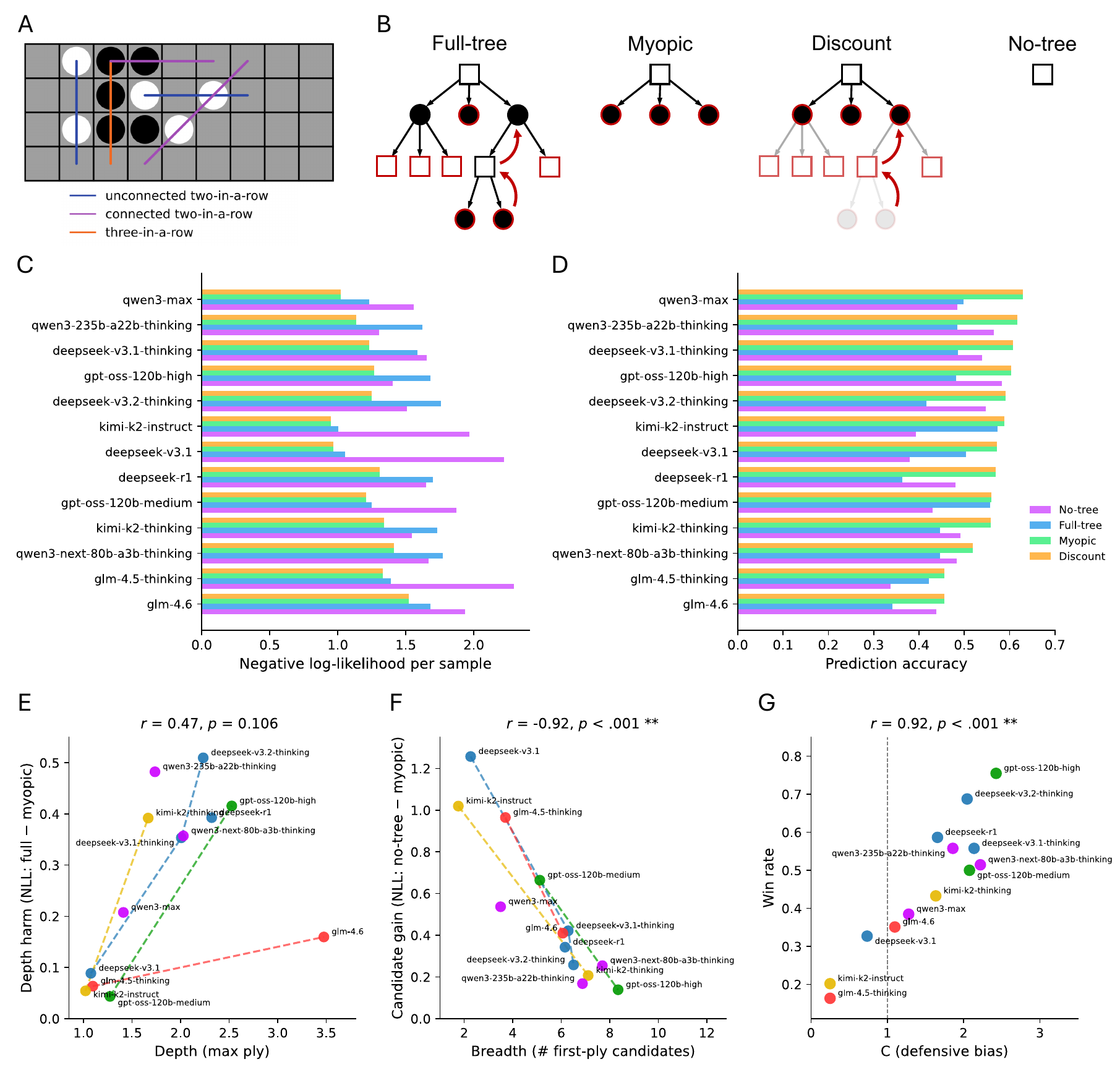}
    \caption{
    \textbf{Predicting moves from extracted search trees with cognitive modeling.}
    (A) Features used in the heuristic value function. Features include connected two-in-a-row, unconnected two-in-a-row, three-in-a-row, a four-in-a-row feature (not shown in the figure), and a center feature. Features with the same type are constrained to have identical weights. 
    (B) Schematics of computational models. The top square represents the current board state. Each downstream circle represents a state resulting from a model’s own simulated move, and each downstream square represents a state resulting from a simulated opponent’s move. Red outlines mark states evaluated by the heuristic function. Red arrows illustrate value backups. In the discount model, deeper nodes are down-weighted (grayed).
    (C) Negative log-likelihood per sample.
    (D) Move prediction accuracy.
    (E) Depth harm, defined as the NLL gap between the full-tree model and the myopic model, as a function of search depth.
    (F) Candidate gain, defined as the NLL gap between the no-tree model and the myopic model, as a function of search breadth.
    (G) Win rate as a function of the fitted active scaling parameter. In the myopic model, this parameter represents the defensive bias. 
    Dashed lines connect models in a model family. Asterisks denote significance levels ($\text{*} \, p < 0.05$, $\text{**} \, p < 0.01$).
    }
    \label{fig:f3}
\end{figure}

\paragraph{Heuristic function}
Following \citet{van2023expertise}, we used a heuristic function $h(s)$ defined as a linear combination of spatial pattern features of the board state $s$:
\begin{equation}
    h(s) = w_\text{center}\bigl(\phi_\text{center}(s,\text{self}) - \phi_\text{center}(s,\text{opp})\bigr) + \sum_{i=1}^4 w_i \bigl(C_\text{self}\,\phi_i(s,\text{self}) - C_\text{opp}\phi_i(s,\text{opp})\bigr),
    \label{eq:heuristic}
\end{equation}
where the center feature $\phi_\text{center}$ sums the inverse Euclidean distance to the board center over a player's pieces, giving higher value to pieces closer to the center. The remaining four features $\phi_i$ count pattern occurrences across all rows, columns, and diagonals. These patterns are connected two-in-a-row, unconnected two-in-a-row, three-in-a-row, and four-in-a-row (\autoref{fig:f3}A). A pattern counts only in windows unobstructed by the opponent, so that a feature reflects a viable threat that can still become four-in-a-row. We allowed the active player’s feature weights (but not the center feature weight) to be scaled by a factor $C$, that is, $C_\text{self}=C$ and $C_\text{opp}=1$ if the model is to move at $s$, and $C_\text{self}=1$ and $C_\text{opp}=C$ if the opponent is to move at $s$. This lets threats be worth more or less to the active player. $C>1$ weights the active player's threats more than the passive player's, and $C<1$ the reverse.

\paragraph{Value backup rules}
Given a search tree, the model computes a value $V(s_i)$ for each depth-1 state $s_i$. We considered four backup rules, corresponding to four model variants (\autoref{fig:f3}B). In the \textit{full-tree} model, the heuristic $h(s)$ is evaluated at all leaf nodes and recursively propagated upward using the minimax rule, with the active player maximizing and its opponent minimizing at alternating plies:
\begin{equation}
    V(s) = \begin{cases}
        h(s) & \text{if } s \text{ is a leaf} \\
        \max_{s'} V(s') & \text{if the model is to move at } s\\
        \min_{s'} V(s') & \text{if the opponent is to move at } s.
    \end{cases}
    \label{eq:minimax}
\end{equation}
Here $s'$ denotes the children of $s$ in the extracted search tree. For the \textit{myopic} model, the heuristic is applied directly to the depth-1 states resulting from each first-ply move, $V(s_i) = h(s_i)$, ignoring everything deeper in the tree. The \textit{discount} model interpolates between these two extremes, where the backed-up value is a weighted sum of the local heuristic and the minimax value of its children:
\begin{equation}
    V(s) = \begin{cases}
        h(s) & \text{if } s \text{ is a leaf} \\
        (1-\gamma)\,h(s) + \gamma\max_{s'} V(s') & \text{if the model is to move at } s \\
        (1-\gamma)\,h(s) + \gamma\min_{s'} V(s') & \text{if the opponent is to move at } s.
    \end{cases}
    \label{eq:discount}
\end{equation}
The free parameter $\gamma \in [0,1]$ controls the influence of deeper nodes. Setting $\gamma = 1$ reduces the discount model to the full-tree model and setting $\gamma = 0$ reduces it to the myopic model. Finally, the \textit{no-tree} model disregards the extracted search tree entirely, scoring every legal first-ply move using the heuristic function alone. This serves as a baseline for testing whether LLMs' search trees carry predictive information beyond the heuristic function.

\paragraph{Model fitting}
Move choice is modeled as a softmax over the backed-up values of candidate moves:
\begin{equation}
    P(s_i \mid s) = \frac{\exp\bigl(V(s_i)\bigr)}{\sum_{s_j \in \mathcal{A}} \exp\bigl(V(s_j)\bigr)},
    \label{eq:choice}
\end{equation}
where $\mathcal{A}$ is the set of depth-1 states resulting from each candidate first-ply move. It includes the depth-1 states in the extracted tree for tree-based models, and all legal depth-1 states for the no-tree model. We did not include a softmax temperature parameter, since it is not separately identifiable from the overall scale of the feature weights and is absorbed into them. Parameters were fitted by minimizing the negative log-likelihood (NLL) over all observed moves using L-BFGS-B \citep{byrd1995limited} with 20 random restarts. We excluded Llama-3.3-70B-Instruct from the following analyses because it produced fewer than 20 parseable search trees, leaving insufficient samples for reliable parameter estimation (see \autoref{appendix:methods}.4 for detailed model fitting methods).

\subsection{LLM moves are best explained by a myopic model}

We compared the four cognitive model variants across all LLMs. Tree-based models consistently outperformed the no-tree baseline, confirming that LLM search trees carry predictive information about move decisions. Among tree-based models, the myopic model achieved lower NLL and higher prediction accuracy than the full-tree model for every LLM (\autoref{fig:f3}C--D; see \autoref{appendix:results}.2 for model recovery analysis and \autoref{appendix:results}.4 for robustness to the choice of heuristic features). On turns where the two models disagreed, the myopic model was uniquely correct more than twice as often (1236 vs.\ 512 turns). This is a striking reversal of the pattern reported by \citet{van2023expertise}, where the tree search model consistently outperformed the myopic model in predicting human moves. We quantified the extent to which the full-tree model underperformed the myopic model as ``depth harm'', defined as the NLL increase from using the full tree over the myopic model. Depth harm was positive for every LLM and tended to increase with search depth (\autoref{fig:f3}E), suggesting that backing up values from deeper nodes consistently impaired rather than improved prediction. The discount model further confirmed this finding: the discount factor $\gamma$ converged to near zero for every LLM, collapsing the model to the myopic model and indicating that deeper nodes received negligible weight in the value backup. Together, these results suggest that LLMs do not perform value backups over the search trees in the way humans do. Although LLMs expand deeper nodes in their reasoning traces, value information from these nodes is not propagated upward to inform first-ply move decisions.

When does access to the search tree improve move prediction? We quantified the gain from having a search tree as ``candidate gain'', defined as the NLL decrease from the no-tree model to the myopic model. Candidate gain decreased strongly with search breadth (\autoref{fig:f3}F). This reveals where the decisive step happens in different LLMs. For narrow-searching LLMs that consider only a few candidates, knowing which moves enter the candidate set is highly predictive of the final move, suggesting that candidate proposal itself is the decisive step. For broad-searching LLMs that consider many candidates, the candidate set barely filters the action space and adds little information beyond the heuristic value function alone. In short, the fewer moves an LLM considers, the more consequential its choice of what to consider becomes. This is consistent with our earlier finding that search breadth, not depth, predicts win rate. What matters for LLM performance is not how deeply a model searches but how broadly it covers the candidate space.

Next, we examined whether the parameters of the myopic model predicted game performance across LLMs. To control for the overall scale of the parameters, we normalized all feature weights relative to the four-in-a-row feature weight. Under this normalization, only the active scaling parameter $C$ significantly predicted win rate (\autoref{fig:f3}G; see \autoref{appendix:results}.3 for all feature weights). Because the myopic model evaluates only the states that follow the LLM's own first-ply move, the opponent is always the active player in these states, so $C$ scales the opponent's threats relative to the LLM's own. Thus, the result suggests that LLMs that weighted their opponent's threats more heavily relative to their own achieved higher performance.

\section{Reasoning traces causally drive move selection, but not through deep search}

We have shown that LLM reasoning traces structurally resemble tree search, and that the relationship between the extracted search tree and move choice is best explained by a model that selects the highest-valued first-ply move. Yet correlation between reasoning structure and move choice does not establish that reasoning causally drives the move decision. The model might reach the same decision regardless of what appears in its trace. To directly test the causal role of reasoning, we conducted a causal intervention analysis in which we surgically removed segments of the reasoning trace, reran inference on the pruned trace, and measured how often the model's final move changed.

\begin{figure}[!t]
    \centering
    \includegraphics[width=\linewidth]{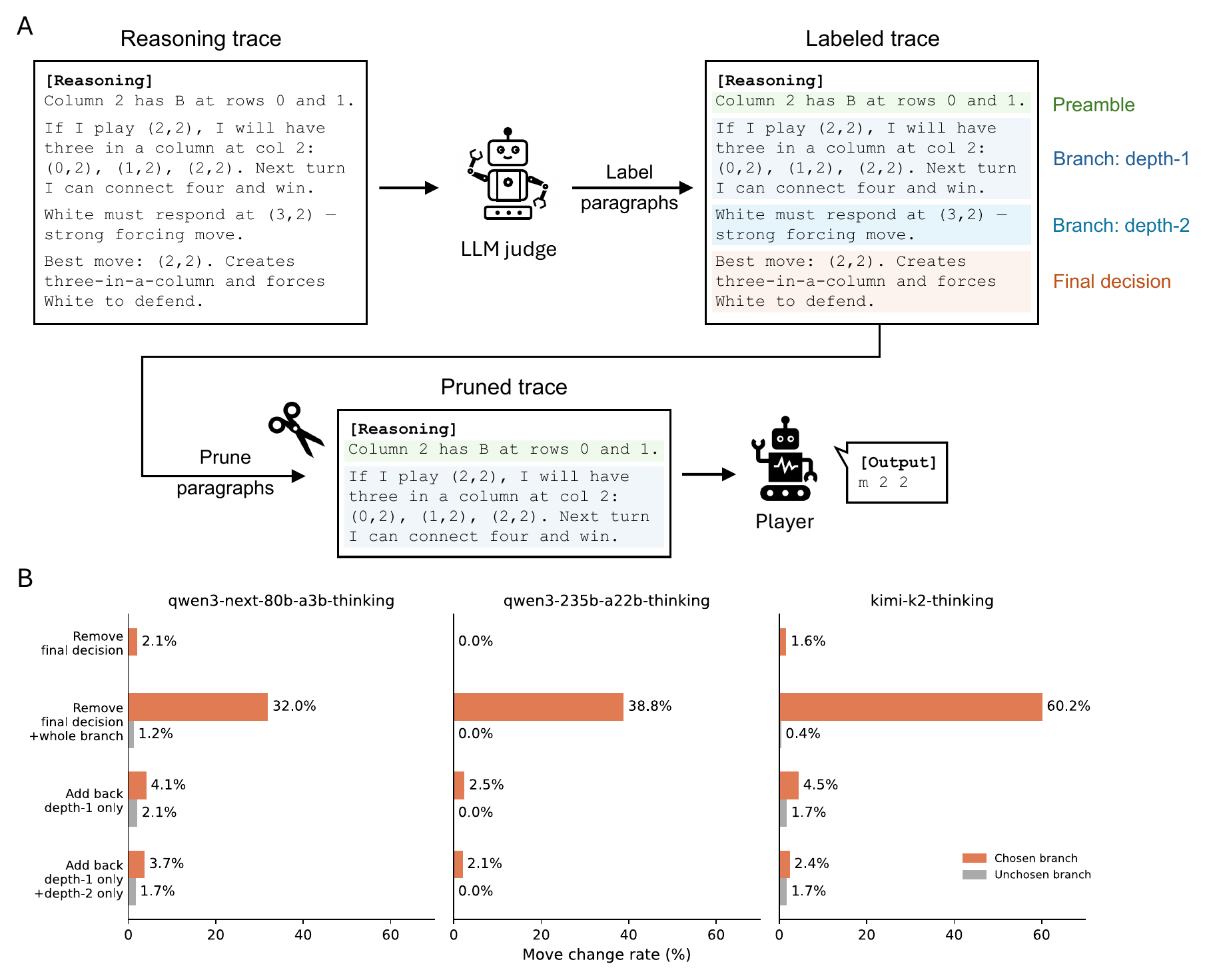}
    \caption{
    \textbf{Causal intervention on reasoning traces.}
    (A) An LLM judge (Claude Opus 4.7) labels each paragraph of the reasoning trace as preamble, branch, final decision, or meta. Branch paragraphs are associated with specific candidate moves. The judge additionally annotates all moves mentioned within each paragraph, together with their search depths. We then prune the trace according to these labels and feed the pruned trace back to the LLM player, which generates a new move.
    (B) Move change rate after four pruning conditions: removing only the final decision, removing both the final decision and an entire branch, and progressively adding back branch paragraphs by depth. Orange: chosen branch; gray: largest unchosen branch.
    }
    \label{fig:f4}
\end{figure}

We performed this analysis on three LLMs (\autoref{fig:f4}). We first labeled each paragraph of the reasoning trace using an LLM judge (Claude Opus 4.7), assigning one of four coarse categories: \emph{preamble} (board parsing and threat scanning before any branch exploration), \emph{branch} (analysis of a specific candidate move and its consequences, corresponding to a branch of the search tree), \emph{final decision} (confirmation of the selected move), or \emph{meta} (general meta-commentary not tied to a specific move, e.g., ``Let me think...'') (\autoref{fig:f4}A; see \autoref{appendix:methods}.5 for detailed intervention methods). The judge also annotated every move evaluated within each paragraph, together with its tree depth. This procedure allowed us to determine whether a paragraph evaluated a future move and which branch of the search tree it belonged to.

We first removed the final-decision paragraphs (\autoref{fig:f4}B). This intervention changed the move in only 0.0\% to 2.1\% of turns, indicating that the final concluding paragraphs did not critically determine the choice. We then additionally removed paragraphs corresponding to a whole branch. Removing the branch corresponding to the chosen move changed the move 32.0\%, 38.8\%, and 60.2\% of the time, respectively, whereas removing the largest unchosen branch changed it only 1.2\%, 0.0\%, and 0.4\% of the time. These results demonstrate that reasoning traces causally drive move selection.

We then identified which content within the chosen branch was causally responsible for the decision. Starting from full branch removal, we progressively added back paragraphs by depth class. Adding back only depth-1 paragraphs---those that mention the candidate move and its immediate consequences---reduced the change rate to 4.1\%, 2.5\%, and 4.5\%, close to the controls in which the same manipulation was applied to the unchosen branch (2.1\%, 0.0\%, and 1.7\%). Adding deeper paragraphs on top reduced the rate by at most 2.1 percentage points (to 3.7\%, 2.1\%, and 2.4\%), indistinguishable from the depth-1 condition and from the corresponding controls (1.7\%, 0.0\%, and 1.7\%), suggesting that deeper lookahead content has negligible causal effect. In other words, even though the models write out deep lookahead in their reasoning, removing that content barely changes their decisions. The shallowest evaluation of each candidate move is what drives the choice.

\section{Discussion}

In this study, we developed a method for extracting and quantifying search trees from LLM reasoning traces during four-in-a-row gameplay and examined how these extracted trees relate to move choices. We found that LLMs search more shallowly than humans, and that their move choices are best explained by a myopic strategy that makes choices based on shallow search while ignoring deep search entirely. These findings point to a fundamental difference between LLM and human planning: while human expertise is driven by deeper lookahead, LLMs do not use deeper search to guide their decisions.

Our findings speak to the ongoing debate about whether LLMs can plan \citep{valmeekam2025systematic, kambhampati2024position}, but suggest that the central question should be reframed. Prior work has largely evaluated planning through behavioral benchmarks. This has led to contested conclusions that depend heavily on task design and prompting strategy \citep{valmeekam2025systematic, sel2025llms, silver2024generalized}. Rather than only asking whether LLMs \emph{succeed} at planning, here we ask what kind of planning \emph{algorithm} their decisions rely on. Answering this question requires looking beyond accuracy and into the internal structure of reasoning. Our approach yields a more fine-grained characterization of LLM planning: LLMs generate the surface structure of tree search, yet their decisions are driven by a myopic mechanism. This dissociation would be invisible to behavioral benchmarks, highlighting the need for mechanistic analyses of reasoning processes to understand planning in LLMs.

More broadly, our results have implications for CoT monitoring and scalable oversight. The dissociation between reasoning traces and move decisions cautions against treating such traces as transparent records of model deliberation. This echoes prior work showing that model-generated intermediate reasoning and explanations can be plausible without being faithful to the model's actual decision process \citep{jacovi2020towards, turpin2023language, lanham2023measuring, paul2024making, korbak2025chain}. If a model can generate structured search without relying on the search to make decisions, then oversight methods based on reasoning traces alone may fail to detect the mechanisms that actually drive behavior. This concern bears directly on the broader challenge of scalable oversight \citep{bowman2022measuring}.

Our findings offer concrete targets for improving and aligning LLM planning. Standard approaches that scale test-time compute, lengthen reasoning traces, or encourage deeper search implicitly assume that additional generated content will shape the model’s final decision. Our results instead suggest that the bottleneck lies not in trace length but in the model's ability to act on what it generates. Closing this gap may therefore require training signals that go beyond outcome supervision and explicitly reward the use of deep lookahead. Such training would also bring LLM planning closer to the depth-driven expertise observed in humans and improve the faithfulness and monitorability of LLM reasoning.

Why are LLMs myopic even when they generate deeper search? Several non-exclusive explanations are possible. First, the bottleneck may be algorithmic. Effective planning requires representing the tree structure and propagating values via Bellman maximization. Transformer attention may fail to implement these operations, especially when the relevant information is distributed across a long context \citep{liu2024lost, paul2024making, lanham2023measuring}. Second, models may learn heuristics other than minimax, such as evaluating salient tactical threats rather than backing up leaf values. Third, myopia may be adaptive under the model's own uncertainty. If predicted game states become increasingly unreliable at greater depths, using them for decisions could hurt more than help \citep{chen2025rational, xiao2019learning, lei2026environmental, jiang2015dependence}. Under this view, shallow reliance is not a failure but a learned policy that treats distant futures as untrustworthy.

The myopia in LLM planning may also reflect a misalignment in computation allocation. Research on human planning has found that people use fast, flat mental simulation when playing a new game for the first time \citep{collins2026people}, while their planning depth grows with expertise \citep{van2023expertise}. LLMs follow neither cleanly. On the one hand, the surface structure of their reasoning traces mirrors human expertise, as if deeper search reflected greater skill. On the other hand, what actually drives their move decisions is flat, shallow evaluation, like that of human novices. One reason for this mismatch may lie in how the two systems allocate computation. For human novices, flat simulation is resource-rational, because it achieves reasonable accuracy at low computational cost \citep{collins2026people}. In contrast, LLMs are known to be cost-insensitive and poor at deploying the right amount of computation adaptively \citep{chen2025not, wang2025thoughts, yang2025towards, de2024rational}, and deliberation that does not inform the decision faces no pressure to be pruned. The dissociation we observe may therefore persist because LLMs lack the cost pressure that, in humans, keeps deliberation coupled to decisions.

Our study has several limitations. First, all analyses were conducted on a single game domain. Whether our conclusions generalize to tasks with different structures and demands remains an open question. Second, we extracted the search trees for LLMs directly from their reasoning traces, whereas \cite{van2023expertise} inferred latent search trees with a cognitive model. Since the two types of search trees were obtained through different procedures, comparisons between LLMs and humans should be interpreted with this difference in mind.

\section*{Acknowledgments and disclosure of funding}

We thank Wei Ji Ma, Jordan Lei, Nastaran Arfaei, Daisy Lin, and Shucheng Li for their helpful comments and feedback on the project. The game simulations were supported by funding from Generality, Inc. The authors declare no competing interests.

\bibliographystyle{plainnat}
\bibliography{references}

\newpage

\appendix

\section{Code and data availability}
\label{appendix:data}

Code is available at \href{https://github.com/generality-inc/fiar-reasoning}{\texttt{https://github.com/generality-inc/fiar-reasoning}}. Raw game logs are available at \href{https://huggingface.co/datasets/SixingChen01/fiar-reasoning}{\texttt{https://huggingface.co/datasets/SixingChen01/fiar-reasoning}}.

\section{List of LLMs used in the study}
\label{appendix:list}

We ran a tournament with 27 models. Models included:
\begin{table}[h]
\centering
\begin{tabular}{l l @{\hspace{1cm}} l l}
\toprule
\textbf{Trace-accessible (14)} & \textbf{API} & \textbf{Summary-only (13)} & \textbf{API} \\
\midrule
DeepSeek-R1-0528            & Fireworks   & GPT-4o          & OpenAI \\
DeepSeek-V3.1               & Vertex AI   & GPT-5 (medium)  & OpenAI \\
DeepSeek-V3.1-Thinking      & Fireworks   & GPT-5 (high)    & OpenAI \\
DeepSeek-V3.2-Thinking      & DeepSeek    & GPT-5 mini (medium) & OpenAI \\
GLM-4.5-Thinking            & Fireworks   & GPT-5 mini (high)   & OpenAI \\
GLM-4.6                     & Zhipu AI    & o3              & OpenAI \\
Kimi-K2-Instruct            & Fireworks   & Claude Sonnet 4     & Vertex AI \\
Kimi-K2-Thinking            & Moonshot    & Claude Sonnet 4.5   & Vertex AI \\
Llama-3.3-70B-Instruct      & Fireworks   & Claude Opus 4.1     & Vertex AI \\
Qwen3-235B-A22B-Thinking    & Fireworks   & Gemini 2.5 Pro  & Google AI \\
Qwen3-Max                   & Alibaba     & Grok-4          & Azure \\
Qwen3-Next-80B-A3B-Thinking & Vertex AI   & Grok-4 Fast Reasoning & Azure  \\
GPT-OSS-120B (medium)       & Vertex AI   & Grok Code Fast 1 & Azure  \\
GPT-OSS-120B (high)         & Vertex AI   &                 & \\
\bottomrule
\end{tabular}
\vspace{0.5em}
\caption{Models evaluated in this study and their APIs.}
\label{tab:models}
\end{table}
\vspace{-15pt}

Note that although Qwen3-Max does not have publicly released weights, its reasoning traces are fully accessible, and we included it in our analyses accordingly.

\section{Additional methods}
\label{appendix:methods}

\newtcolorbox{promptbox}[1]{
  enhanced,
  breakable,
  colback=gray!5,
  colframe=gray!5,
  coltitle=white,
  colbacktitle=black!85,
  fonttitle=\bfseries,
  title=#1,
  sharp corners=south,
  boxrule=0.4pt,
  left=6pt, right=6pt, top=6pt, bottom=6pt,
}

\subsection{Game tournament and trace collection}

Models played four-in-a-row on a 4$\times$9 board against each other via an automated game loop. At each turn the model received a fixed system prompt defining the game rules and move notation, followed by a user message containing the current board state in FEN notation and the current player. Models with extended thinking produced both a reasoning trace and a final move answer wrapped in tags. We retained both for downstream analyses.

\begin{promptbox}{System Prompt: Four-in-a-row Game}
Let's play a game of Four in a Row. You are playing as [White/Black]. You will be given the current game state and you will need to give the next move in a standard algebraic notation specific to this game. Feel free to think about the move, only the final answer you provide in \texttt{<next\_move> </next\_move>} tags will be played.

\medskip
\textbf{Game Rules:}\\
Four in a row is played on a four-by-nine grid by two players, who alternately place the marks W and B in one of the thirty-six spaces in the grid. A player wins when they get four pieces in a row horizontally, vertically or diagonally. Player W plays first.

\medskip
\textbf{The standard game state representation is in the following format:}\\
The game state will be represented in FEN notation, a compact algebraic representation inspired by chess's Forsyth-Edwards Notation. Each row of the 4$\times$9 board is encoded as a string where `W' represents a White piece, `B' represents a Black piece, and numbers indicate consecutive empty spaces. Rows are separated by forward slashes (`/'), reading from top to bottom. An empty board is represented as \texttt{9/9/9/9}, with each `9' indicating that all nine columns in that row are empty.

\medskip
\textbf{Standard Algebraic Notation (SAN) Explanation:}\\
Issue moves in the notation \texttt{m <row> <col>}, for example \texttt{m 0 0} to place your mark in the top leftmost square and \texttt{m 3 8} to place your mark in the bottom rightmost square.
\end{promptbox}

\begin{promptbox}{User Prompt: Four-in-a-row Game}
The current board state is:\\
FEN: \{fen string\}

\medskip
Current player: \{White/Black\} (\{W/B\})
\end{promptbox}

\subsection{Search tree extraction}

Reasoning traces are free-form natural language and cannot be reliably parsed by rule-based methods. We used GPT-5 (medium reasoning effort) as a judge. Given a reasoning trace, it reconstructed the set of moves the model explicitly considered as a nested JSON array. Each node in the array is a list whose first element is a coordinate string \texttt{"r,c"} and whose remaining elements are children representing replies.

The system prompt for this step was automatically optimized using DSPy \citep{khattab2024dspy} with the GEPA optimizer \citep{agrawal2026gepa}. Starting from a hand-written specification of the extraction task, the optimizer iteratively proposed and evaluated candidate instruction sets against a human-annotated set, selecting the instruction text that maximized extraction accuracy. The resulting optimized prompt is reproduced in full below.

\begin{promptbox}{System Prompt: Search Tree Extraction}
You are given a natural-language reasoning trace about a 4$\times$9 Four-in-a-Row (Connect-X-like) game. Your task is to reconstruct exactly the explicitly considered move trees and output strict JSON in a prescribed nested array format.

\medskip
\textbf{Domain and board facts} (authoritative; use as-is, do not infer/alter):
\begin{itemize}
  \item Board: 4 rows $\times$ 9 columns, zero-based. Rows 0--3 (top to bottom), columns 0--8 (left to right).
  \item No gravity: a move may be played in any empty cell. Ignore any ``falling piece'' assumptions.
  \item Win: four in a row horizontally, vertically, or diagonally (both down-right and down-left).
  \item Vertical four is exactly a whole column (rows 0--3 in that column).
  \item Diagonals of length 4 are exactly:
    \begin{itemize}
      \item Down-right (top-left $\to$ bottom-right): start only at row 0, columns 0--5. Each is \texttt{(0,c),(1,c+1),(2,c+2),(3,c+3)}.
      \item Down-left (top-right $\to$ bottom-left): start only at row 0, columns 3--8. Each is \texttt{(0,c),(1,c-1),(2,c-2),(3,c-3)}.
    \end{itemize}
\end{itemize}

\medskip
\textbf{Accepted coordinate inputs and normalization:}
\begin{itemize}
  \item Accept these explicit coordinate forms:
    \begin{itemize}
      \item \texttt{"(r,c)"}
      \item \texttt{"row r col c"}
      \item \texttt{"m r c"} (move notation; treat exactly as \texttt{r,c})
    \end{itemize}
  \item Accept explicit enumerations and expand to individual coordinates:
    \begin{itemize}
      \item \texttt{"row 2 cols 6,7,8"} $\to$ add \texttt{"2,6"}, \texttt{"2,7"}, \texttt{"2,8"}
      \item \texttt{"column 5 rows 0,1,3"} $\to$ add \texttt{"0,5"}, \texttt{"1,5"}, \texttt{"3,5"}
      \item \texttt{"row 1 col 2"} (single), \texttt{"row 1 col 1 or col 0"} $\to$ add \texttt{"1,1"} and \texttt{"1,0"}
    \end{itemize}
  \item Normalize every coordinate to the exact string \texttt{"r,c"} (no spaces).
  \item Unambiguous-only: Ignore vague phrases (e.g., ``left side'', ``near the center'', ``the other end'') unless the exact endpoint squares were enumerated earlier in that same branch context.
\end{itemize}

\medskip
\textbf{Required output format} (strict JSON):
\begin{itemize}
  \item Produce exactly: \texttt{\{"trees": Node[]\}}
  \item \texttt{Node := ["r,c", Node, Node, ...]}
    \begin{itemize}
      \item First element is the move coordinate string \texttt{"r,c"}.
      \item Following elements (if any) are child Node values.
    \end{itemize}
  \item A leaf node is just \texttt{["r,c"]}.
  \item Strict JSON only:
    \begin{itemize}
      \item Use double quotes for all strings.
      \item No comments, no trailing commas, no extra fields.
      \item Do not wrap your final JSON in Markdown/code fences.
    \end{itemize}
\end{itemize}

\medskip
\textbf{Tree semantics and strict alternation:}
\begin{itemize}
  \item Each depth-1 node is a model-side first move explicitly considered in the trace.
  \item Depth alternates by side:
    \begin{itemize}
      \item Depth 1: model move
      \item Depth 2: opponent reply
      \item Depth 3: model reply
      \item Depth 4: opponent reply
      \item \ldots{} and so on. Maintain alternation at every depth.
    \end{itemize}
  \item Children under any node are exactly the explicitly described replies for that exact position/branch.
\end{itemize}

\medskip
\textbf{How to extract moves from the trace} (precise, comprehensive):

\begin{enumerate}
\item \textbf{Identify model-side root moves (first-move candidates only).}
  \begin{itemize}
    \item Create a separate node for every distinct coordinate the trace explicitly considers as the model's first move.
    \item Treat as model-first-move candidates when phrased as:
      \begin{itemize}
        \item ``If I play (r,c)\ldots'', ``I can/could/should play (r,c)\ldots'', ``What if I play (r,c)\ldots''
        \item ``Another idea: (r,c)\ldots'', ``The best move is (r,c)\ldots'' (final chosen move still counts as a node)
        \item Enumerations: ``I could play (a,b) or (c,d)'' $\to$ add both as separate node.
        \item Series: ``I could fill (2,1),(2,2),(2,3)\ldots'' $\to$ add each as a separate node.
        \item \texttt{"m r c"} used to propose the model's move choice $\to$ add as a node.
      \end{itemize}
    \item Include every distinct explicitly proposed model-first-move, even if later rejected or deemed inferior. Do not drop quick tests or briefly examined options.
    \item Do NOT promote coordinates if they appear only as opponent moves or only as deeper replies (unless the trace also proposes them as a model-first move elsewhere).
  \end{itemize}

\item \textbf{Build subtrees per depth-1 node (no cross-pollination across nodes).}
  \begin{itemize}
    \item Under a specific depth-1 node, attach opponent replies tied to that exact node scenario. Use phrasing such as: ``then they can\ldots'', ``the opponent can reply at\ldots'', ``they must block at\ldots'', ``they'll answer with\ldots'', ``they can try\ldots'', ``they target\ldots''.
    \item Include opponent move enumerations tied to that node, in the order explicitly mentioned:
      \begin{itemize}
        \item Column/row enumerations: e.g., ``Their vertical in column 5 needs rows 0,1,3'' $\to$ add \texttt{"0,5"},\texttt{"1,5"},\texttt{"3,5"}.
        \item Horizontal targets: ``After (0,3), they threaten row 2 cols 6,7,8'' $\to$ add \texttt{"2,6"},\texttt{"2,7"},\texttt{"2,8"}.
        \item Diagonal endpoints: If a diagonal is described or implied by endpoints (e.g., ``(0,3)-(1,4)-(2,5)-(3,6)'' or ``they could play (0,3) or (3,6)''), add each explicitly named endpoint as an opponent child.
      \end{itemize}
    \item For each opponent child, attach exactly the model's explicitly stated reply(ies) for that branch, maintaining alternation. If multiple model replies are given (``I must play (2,2) or (2,6)''), add both as children under that opponent node.
    \item Continue deeper when the trace specifies the next opponent move after the model reply (e.g., ``after I block at (2,2), they win by the other end (2,6)''), adding that opponent move as the next child node at the correct depth.
    \item Only include moves explicitly tied to the current context. Do NOT transfer opponent replies or deeper sequences from one node to another unless the trace explicitly restates them for that other node.
  \end{itemize}

\item \textbf{Mirroring, ``if instead/similarly'', and ``the other end'' within the same node.}
  \begin{itemize}
    \item If the trace presents mirrored alternatives (``or instead at (c,d)'', ``similarly at (x,y)''), include each as a sibling child under the same parent, in order of mention.
    \item ``The other end'' is allowed only if both endpoints of the line were explicitly established earlier in that branch. Add the mirrored endpoint if explicitly identified or implied by previously listed endpoints of that same line.
    \item When both ends of a horizontal or diagonal threat are explicitly named, include both as children.
  \end{itemize}

\item \textbf{Context tracking and deeper sequences.}
  \begin{itemize}
    \item Maintain the exact branch context as described. When the trace walks through a sequence (``If I play (0,1), they must play (3,4); then I play (0,5); then they must block (3,2); then I can play (2,5)\ldots''), build the path depth-by-depth under the same depth-1 node with strict alternation.
    \item If the narrative continues with ``after that'', ``then'', ``from there'', or refers back to a previously described branch (without switching to a new branch), continue that exact subtree.
    \item Do not assume or create branches not explicitly stated.
  \end{itemize}

\item \textbf{Unambiguous-only policy.}
  \begin{itemize}
    \item Include only coordinates that are explicitly specified (via accepted formats or explicit enumerations).
    \item Ignore vague references (``block there'', ``left side'', ``near (r,c)'') unless the exact squares were enumerated earlier in that same branch.
  \end{itemize}

\item \textbf{De-duplication and ordering.}
  \begin{itemize}
    \item Within the same parent, include each child coordinate at most once.
    \item If a node reappears later under the same parent, merge newly stated children into that existing node (preserve alternation).
    \item Preserve the order of children exactly as they are mentioned in the trace.
  \end{itemize}

\item \textbf{Interpreting \texttt{"m r c"} notation.}
  \begin{itemize}
    \item Treat \texttt{"m r c"} exactly as the explicit coordinate \texttt{(r,c)} and normalize to \texttt{"r,c"}.
    \item Add it as a depth-1 node only if used to propose the model's first move (including the final chosen move).
    \item When \texttt{"m r c"} appears as an opponent reply in-branch, add it at the appropriate depth under the correct parent.
  \end{itemize}
\end{enumerate}

\medskip
\textbf{Generalizable extraction strategy} (to avoid common mistakes seen in prior attempts):
\begin{itemize}
  \item Be exhaustive in capturing nodes: every time the narrator explicitly proposes a candidate for ``my first move'' (including quick tests, alternatives, and series), create a node for that coordinate. Do not skip less-preferred or rejected first-move candidates.
  \item Under each depth-1 node, be exhaustive in capturing opponent replies explicitly tied to that node, including:
    \begin{itemize}
      \item All row/column enumerations, expanded to individual squares.
      \item Both explicitly named endpoints of diagonals/horizontals.
      \item Any ``they could play/try/need/must'' squares enumerated as part of threats on that branch.
    \end{itemize}
  \item Keep branches separate by nodes; do not mix children across nodes.
  \item Maintain strict side alternation at all depths.
  \item Do not invent nodes; only add squares explicitly named or enumerated under the current branch context.
  \item Do not promote opponent-only or deeper-branch-only squares to depth-1 nodes unless they were also explicitly presented as model-first-move options elsewhere in the trace.
\end{itemize}

\medskip
\textbf{Final validation checklist before output:}
\begin{itemize}
  \item Top-level object is exactly \texttt{\{"trees": [...]\}} with \texttt{Node[]} content.
  \item Every node is \texttt{["r,c", ...children...]} with coordinates normalized to \texttt{"r,c"} (no spaces).
  \item Depth-1 nodes include every distinct explicit model-first-move candidate from the trace (including all enumerated candidates and quick tests), in order of first mention.
  \item Under each depth-1 node, all explicitly stated opponent replies for that node are present (including all expanded row/column enumerations and diagonal/horizontal endpoints), in order of first mention.
  \item Under each opponent node, all explicitly stated model replies are present; include deeper opponent follow-ups when specified.
  \item No duplicates at the same parent; merge re-mentioned nodes' children; preserve mention order.
  \item Strict alternation by depth is preserved everywhere.
  \item Strict JSON only; use double quotes; no trailing commas; no extra fields; do not wrap in Markdown.
\end{itemize}
\end{promptbox}

\begin{promptbox}{User Prompt: Search Tree Extraction}
\texttt{[[ \#\# trace \#\# ]]}\\
\{reasoning trace\}

\medskip
Respond with a JSON object in the following order of fields: \texttt{trees} (must be formatted as a valid Python \texttt{list[list[Any]]}).
\end{promptbox}

\subsection{Validation of the tree extractor}

\subsubsection{Validation against a human-annotated validation set}

We built a human-annotated set of reasoning traces and validated the tree extractor against the set. We randomly sampled 140 reasoning traces with 10 from each of the 14 trace-accessible models. This set was held out from the set used to optimize the extraction prompt. The annotator read each full trace and transcribed the search tree by hand. This gave one human tree per trace to compare against the judge's output.

To quantify agreement, we used a validation metric based on tree-edit distance \citep{zeng2026rejump}. The metric is tree similarity, $\text{Sim}_T = 1 - \text{TED}(T,T') / \max(|T|, |T'|)$, where $\text{TED}$ is the tree-edit distance between the two search trees. The similarity is normalized to $[0,1]$, with 1 indicating identical trees. Because the candidate moves form an unordered set, we used the order-invariant constrained tree-edit distance \citep{zhang1996constrained}. The judge reached $\text{Sim}_T = 0.956$ against the human annotations, comparable with previous work using a similar method on different tasks and tree semantics \citep{zeng2026rejump}.

We also report strict per-trace agreement between the human annotator and the judge. Here, a trace matches on a dimension only if the extracted tree agrees exactly between the human annotator and the judge, and one misplaced node fails the whole trace. Across the 140 traces, the judge matched the human tree on max ply in 98.6\% of traces, breadth in 87.9\%, edges in 85.0\%, node set in 84.3\%, and the whole tree in 75.0\%. These whole-tree failures were notably small: the median mismatch differed from the human tree by a single node, with the larger disagreements confined to the longest traces (40--60 nodes). Together, these results show that the judge extracted trees reliably.

To examine whether the extractor had a systematic bias, we computed per-node recall and precision at each depth on the 140 traces. Recall at a given depth is the fraction of human-annotated nodes at that depth that the judge recovers, and precision is the fraction of the judge’s extracted nodes at that depth that appear in the human tree. Low recall would indicate omitted nodes, while low precision would indicate adding spurious nodes. Recall declined mildly with depth (0.973, 0.930, 0.895, 0.837 at depths 1–4), and precision declined comparably (0.977, 0.925, 0.951, 0.837). Extracted node counts closely matched the human counts at every depth (depth 1: 796 vs. 800; depth 2: 187 vs. 186; depth 3+: 158 vs. 164). The judge therefore disagreed with the annotator symmetrically about which nodes were present and did not systematically omit or over-extract deeper nodes.

\subsubsection{Validation against a heuristic process}

As an independent verification, we additionally built a parser following a heuristic rule. The parser detects the three coordinate forms ("(r,c)", "row r col c", "m r c") and uses them to cross-check the coordinates the LLM judge extracted. Against the human annotation, the judge and the parser each recovered 97\% of the marked coordinates. This is an independent confirmation that the judge did not miss moves mentioned in the trace.

Nevertheless, the parser alone cannot serve as full ground-truth extraction for two reasons. First, it recovers coordinates but not the tree structure, so it cannot reproduce the branching and depth our analysis relies on. Second, LLMs also use reasoning traces to parse the board, and this board parsing mentions the same coordinates as deliberated moves. A regex thus cannot distinguish between board parsing and deliberation. For example, to parse the board, a model may write "White has a piece at (1,3)." A regex would record a move even though no move is being simulated. Consequently, the heuristic parser extracted 18.1 coordinates per trace, compared with 6.3 human-annotated moves.

\subsection{Computational cognitive model fitting}

\subsubsection{Data exclusion criteria for model fitting}

We applied the following exclusion criteria to the raw data before model fitting:

\begin{itemize}[leftmargin=*]
    \item \textbf{Invalid move}: turns where the model failed to produce a valid move or where the extracted move did not conform to the expected format were excluded.
    \item \textbf{No search tree}: turns where the model produced no search tree were excluded, as these provide no information about the model's search process.
    \item \textbf{Degenerate tree}: turns where the tree contained fewer than two candidate first-ply moves were excluded, as these leave no meaningful choice to model.
    \item \textbf{Chosen move not in tree}: turns where the model's chosen move did not appear among the candidate first-ply moves in the tree were excluded.
    \item \textbf{Insufficient samples}: models for which fewer than 20 turns remained after turn-level exclusions were excluded as insufficient for reliable parameter estimation. This criterion excluded Llama-3.3-70B-Instruct from the analysis.
\end{itemize}

\subsubsection{Parameter ranges of computational models}

We fit computational models to each LLM's move choices. Parameters were estimated by minimizing the negative log-likelihood using L-BFGS-B with 20 random restarts to mitigate local minima. To enforce parameter constraints, scaling factor $C$ was reparameterized as $\exp(\log C)$ and bounded to $C \in [0.25, 5.00]$. Discount factor $\gamma$ was bounded to $\gamma \in [0, 1]$. Feature weights were unconstrained.

\subsection{Causal intervention study}

\subsubsection{Paragraph-level trace labeling}

To surgically remove a branch, we must first identify which paragraphs of the trace structurally belong to it. We split each trace into paragraphs at double newline boundaries and called Claude Opus 4.7 to assign three annotations to every paragraph. These annotations include a structural type, a branch root, and a mentions list recording the move coordinates explicitly simulated within that paragraph along with their lookahead depth.

The structural type captures the paragraph's role in the reasoning process. It takes one of seven values. \texttt{PREAMBLE} paragraphs describe the board state and scan for threats before any candidate move is introduced. \texttt{BRANCH\_START} marks the first paragraph that explicitly proposes a specific move as a candidate (``If I play X\ldots'', ``Let me try X\ldots''). \texttt{BRANCH\_ANALYSIS} and \texttt{BRANCH\_CONCLUSION} label subsequent paragraphs that continue and close the analysis within that branch (opponent replies, counter-replies, local evaluations). \texttt{COMPARISON} marks cross-branch paragraphs that weigh multiple candidate moves against each other. \texttt{FINAL\_DECISION} marks the paragraph(s) that confirm the model's chosen move (``I'll play X'', ``Therefore X''). Finally, \texttt{META} captures generic meta-commentary not tied to any move (``Let me think step by step\ldots'').

In addition to the structural type, each paragraph receives two further annotations. The first is a branch root, recording which candidate first-ply move the paragraph structurally belongs to. The second is a mention list, recording every coordinate that the paragraph explicitly simulates as a future move, each tagged with its depth in the lookahead sequence (depth 1 corresponds to the model's move, depth 2 corresponds to the opponent's reply, depth 3 corresponds to the model's counter-reply, and so on). The mention list only includes prospective moves and excludes coordinates that merely describe and parse the current board occupancy.

\begin{promptbox}{System Prompt: Reasoning Trace Labeling}
You are assisting a research project studying how chain-of-thought reasoning influences LLM behavior. The goal is to surgically remove specific reasoning branches from a model's thinking trace and then re-run the model with the edited trace as a prefill --- observing whether the model changes its decision when it can no longer ``see'' the reasoning that led to a particular move. Accurate labeling is critical: if the wrong paragraphs are removed the intervention is invalid; if key paragraphs (especially final confirmations) are missed the model will trivially repeat the same answer from the residual context.

You are analyzing a reasoning trace from a Four-in-a-Row game player (4 rows $\times$ 9 columns, zero-indexed). Coordinates appear as \texttt{(r,c)}, \texttt{row r col c}, or \texttt{m r c}.

The trace has been split into numbered paragraphs. Your task: assign a label to EVERY paragraph so that we can later remove the branch for a specific target move.

\medskip
\textbf{Label definitions}\\
\texttt{type} --- one of:
\begin{itemize}
  \item[] \texttt{PREAMBLE}: Board parsing, board state description, threat scanning before any branch exploration
  \item[] \texttt{BRANCH\_START}: First paragraph proposing a specific move as the model's candidate first move (``If I play X\ldots'', ``Let me try X\ldots'', ``What if I place at X\ldots'')
  \item[] \texttt{BRANCH\_ANALYSIS}: Continuation of analysis within a branch (opponent replies, counter-replies, evaluations)
  \item[] \texttt{BRANCH\_CONCLUSION}: Local verdict closing a branch (``X looks strong/weak because\ldots'')
  \item[] \texttt{COMPARISON}: Cross-branch comparison mentioning multiple candidate moves
  \item[] \texttt{FINAL\_DECISION}: Confirmation of the chosen move (``I'll play X'', ``My move is X'', ``Therefore X'')
  \item[] \texttt{META}: Meta-commentary not tied to a specific move (``Let me think\ldots'', ``First I'll consider\ldots'')
\end{itemize}

\medskip
\texttt{branch\_root} --- the depth-1 candidate move this paragraph structurally belongs to, as \texttt{"r,c"}. Use \texttt{null} for \texttt{PREAMBLE}, \texttt{COMPARISON}, \texttt{FINAL\_DECISION}, \texttt{META}. Important: assign the same \texttt{branch\_root} to ALL paragraphs in a branch, even strategic preamble paragraphs that motivate the move before naming it.

\medskip
\texttt{mentions} --- list of coordinates that are part of MOVE SIMULATION only, each with:
\begin{itemize}
  \item[] \texttt{coord} --- \texttt{"r,c"}
  \item[] \texttt{depth} --- 1=model's move, 2=opponent reply, 3=model counter-reply, etc.
\end{itemize}

IMPORTANT: only include a coordinate if it is being proposed or simulated as a future move (e.g.\ ``if I play X'', ``opponent responds at Y'', ``let me try X''). Do NOT include coordinates that merely describe the current board state (e.g.\ ``there are pieces at X and Y'', ``X is occupied'', ``the board shows pieces at X, Y, Z''). Board-parsing references are not move simulations and must be excluded from \texttt{mentions}.

\medskip
\textbf{Output format}\\
Return a JSON array with one object per paragraph (in order):
\begin{verbatim}
[
  {"para": 0, "type": "PREAMBLE", "branch_root": null, "mentions": []},
  {"para": 1, "type": "BRANCH_START", "branch_root": "1,4",
   "mentions": [{"coord": "1,4", "depth": 1}]},
  ...
]
\end{verbatim}
Return ONLY the JSON array. No explanation, no markdown fences.
\end{promptbox}

\begin{promptbox}{User Prompt: Reasoning Trace Labeling}
The model is playing as \{color\}. So `\{color\} plays X' = depth 1 (model's move), `\{opponent\} plays Y' = depth 2 (opponent reply).

\medskip
Search tree (depth-1 = model's candidate first moves):\\
\texttt{\ \ root=\{r\},\{c\} (\{N\} nodes)}\\
\texttt{\ \ \ldots}

\medskip
Numbered paragraphs of the reasoning trace:
\begin{verbatim}
<trace>
[0] paragraph 0 text ...

[1] paragraph 1 text ...

[2] paragraph 2 text ...
...
</trace>
\end{verbatim}

Label every paragraph with its type, \texttt{branch\_root}, and \texttt{mentions}. Return a JSON array with one object per paragraph.
\end{promptbox}

\subsubsection{Validation of the judge}

To validate the judge, we compared its labels against a human-annotated validation set. We sampled 150 paragraphs with 50 from each of the three reasoning models we analyzed, and annotated each paragraph by hand. The human annotator and the judge agreed on the final decision paragraphs for all 150 paragraphs. On branch paragraphs, the annotator identified 109 move mentions; the judge recovered 98.2\% of these, and 97.2\% of the recovered moves were assigned the same depth. The judge thus labeled paragraphs reliably.

\subsubsection{Trace editing}

We applied four editing strategies to isolate which parts of a reasoning trace causally drive move selection. Across all strategies, \texttt{FINAL\_DECISION} paragraphs were always removed so the model had to re-decide from the remaining context rather than simply echoing its prior conclusion. The strategies differed in how much additional content was removed on top of this.

\begin{itemize}[leftmargin=*]
    \item \textbf{Remove final decision} removed only \texttt{FINAL\_DECISION} paragraphs, serving as a minimal baseline.
    
    \item \textbf{Remove final decision + whole branch} additionally removed the entire reasoning branch for the chosen move: all paragraphs whose branch root equals the target move, as well as any \texttt{COMPARISON} paragraphs in which the target move appears in the mention list or the paragraph text. This was the maximal intervention.
    
    \item \textbf{Add back depth-1 only} started from the maximal intervention and restored paragraphs whose mention list contains exclusively first-ply moves (i.e., the model's own moves).
    
    \item \textbf{Add back depth-1 only + depth-2 only} further restored paragraphs whose mention list contains exclusively second-ply moves (i.e., the opponent's replies), in addition to the depth-1 paragraphs restored above.
\end{itemize}

In all strategies, we merged contiguous removed spans, collapsed trailing whitespace, and then re-joined the trace. As a safeguard, any edit that would remove more than 85\% of the original trace was discarded to avoid degenerately short traces.

\subsubsection{Model re-running with edited trace}

The edited trace was used as a reasoning prefill. We reconstructed the original prompt, appended the model's thinking-start token, injected the edited reasoning trace, closed the thinking block, and ran the model at temperature~0 with a maximum of 64 new output tokens. The model was required to generate its move in the standard tags. We recorded the model's new move and examined whether it differed from its original move.

\section{Additional results}
\label{appendix:results}

\setcounter{figure}{0}
\renewcommand{\thefigure}{S\arabic{figure}}

\subsection{Search tree depth}

\begin{figure}[h]
    \centering
    \includegraphics[width=\linewidth]{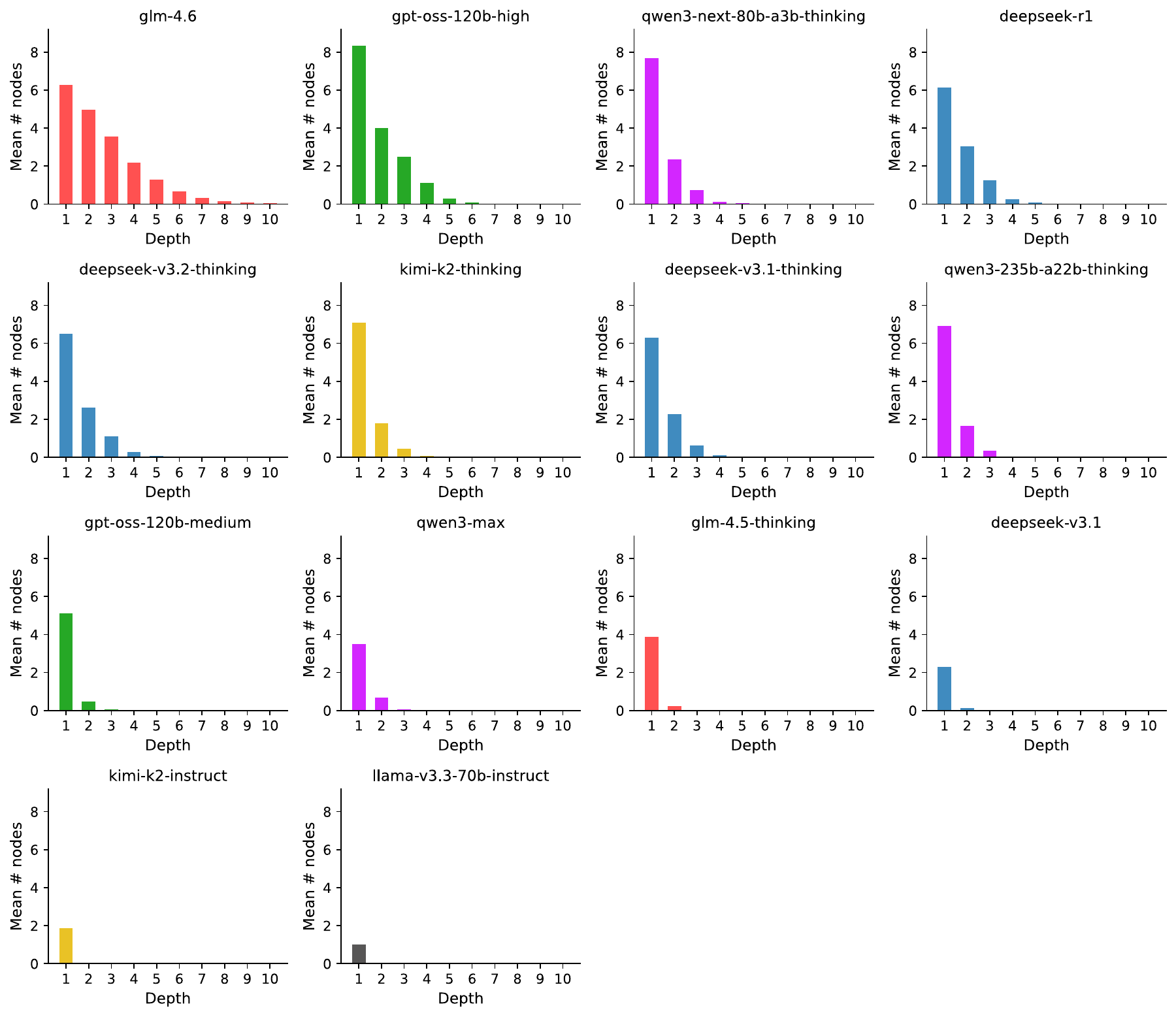}
    \caption{
    \textbf{Distribution of node depth} for each LLM.}
    \label{fig:sf1}
\end{figure}

\autoref{fig:sf1} shows the average number of nodes at each depth of the search tree for each LLM. Depth 1 corresponds to the model's own candidate first moves, depth 2 to the opponent's replies, depth 3 to the model's counter-replies, and so on. All models peaked sharply at depth 1, reflecting that candidate first-ply moves were always the most numerous nodes. The depth-2 count was consistently lower, indicating that models considered fewer opponent replies per candidate first-ply move. Beyond depth 2, node count dropped rapidly for most models. There was nevertheless substantial variation across models in how far ahead they looked.

\subsection{Model recovery analysis}

We contrasted two competing backup rules: a full-tree model that backs up values via minimax across the whole search tree, and a myopic model that evaluates only the immediate position after each candidate first-ply move. A natural concern is whether the fitting procedure can reliably distinguish between these two data-generating mechanisms. A good fit by the myopic model, for instance, could potentially be an artifact of the optimizer finding spurious parameter configurations. To address this concern, we conducted a model recovery analysis.

For each of the 13 LLMs with sufficient data, we ran a two-condition model recovery analysis using the model's own extracted trees:

\begin{enumerate}[leftmargin=*]
    \item \textbf{Simulate from the full-tree model.} Using the model's fitted full-tree parameters, we sampled synthetic move choices from the full-tree softmax policy. We then fit both the full-tree model and the myopic model to these synthetic choices. If the fitting procedure is valid, the full-tree model should outperform the myopic model ($\Delta > 0$, where $\Delta = (\text{NLL}_\text{myopic} - \text{NLL}_\text{full}) / N$).

    \item \textbf{Simulate from the myopic model.} Using the model's fitted myopic parameters, we sampled synthetic move choices from the myopic softmax policy. We then fit both models to these synthetic choices. If the fitting procedure is valid, the myopic model should outperform the full-tree model ($\Delta < 0$).
\end{enumerate}
LLMs with both $\Delta > 0$ in the first condition and $\Delta < 0$ in the second condition were counted as successfully recovered.

\begin{figure}[h]
    \centering
    \includegraphics[width=\linewidth]{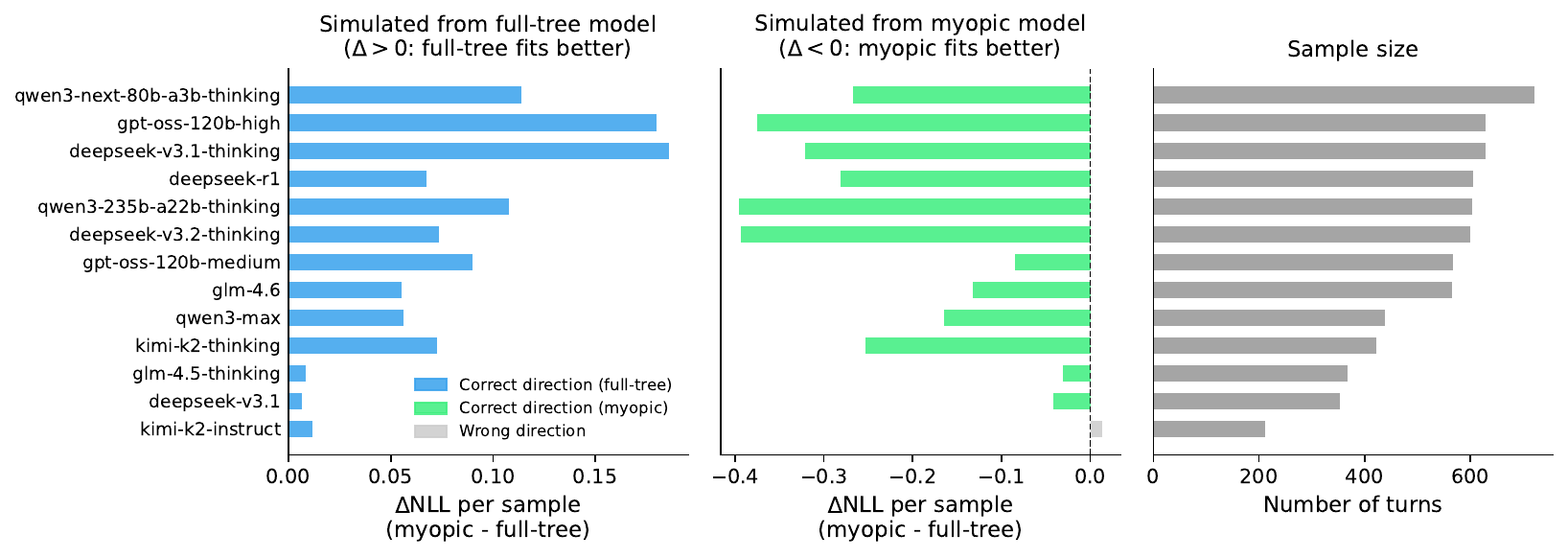}
    \caption{
    \textbf{Model recovery analysis} shows successful recovery of underlying ground-truth models.}
    \label{fig:sf2}
\end{figure}

Model recovery succeeded in 12 out of 13 LLMs (\autoref{fig:sf2}). In the left panel, all 13 LLMs showed positive $\Delta$ when data were generated from the full-tree model, confirming the fitting procedure correctly identifies the full-tree model as superior. In the right panel, 12 of 13 LLMs showed negative $\Delta$ when data were generated from the myopic model. The single failed LLM was Kimi-K2-Instruct, for which the myopic model failed to outperform the full-tree model on its own simulated data. This LLM had the smallest sample in the dataset ($N = 211$), so the signal was insufficient to overcome noise. Notably, all other LLMs recovered successfully in both directions.

These results establish that our fitting procedure can reliably distinguish full-tree from myopic decision-making given sufficient data. The model comparisons are therefore not artifacts of optimizer behavior or parameter degeneracy.

\subsection{Fitted feature weights}

\begin{figure}[h]
    \centering
    \includegraphics[width=0.667\linewidth]{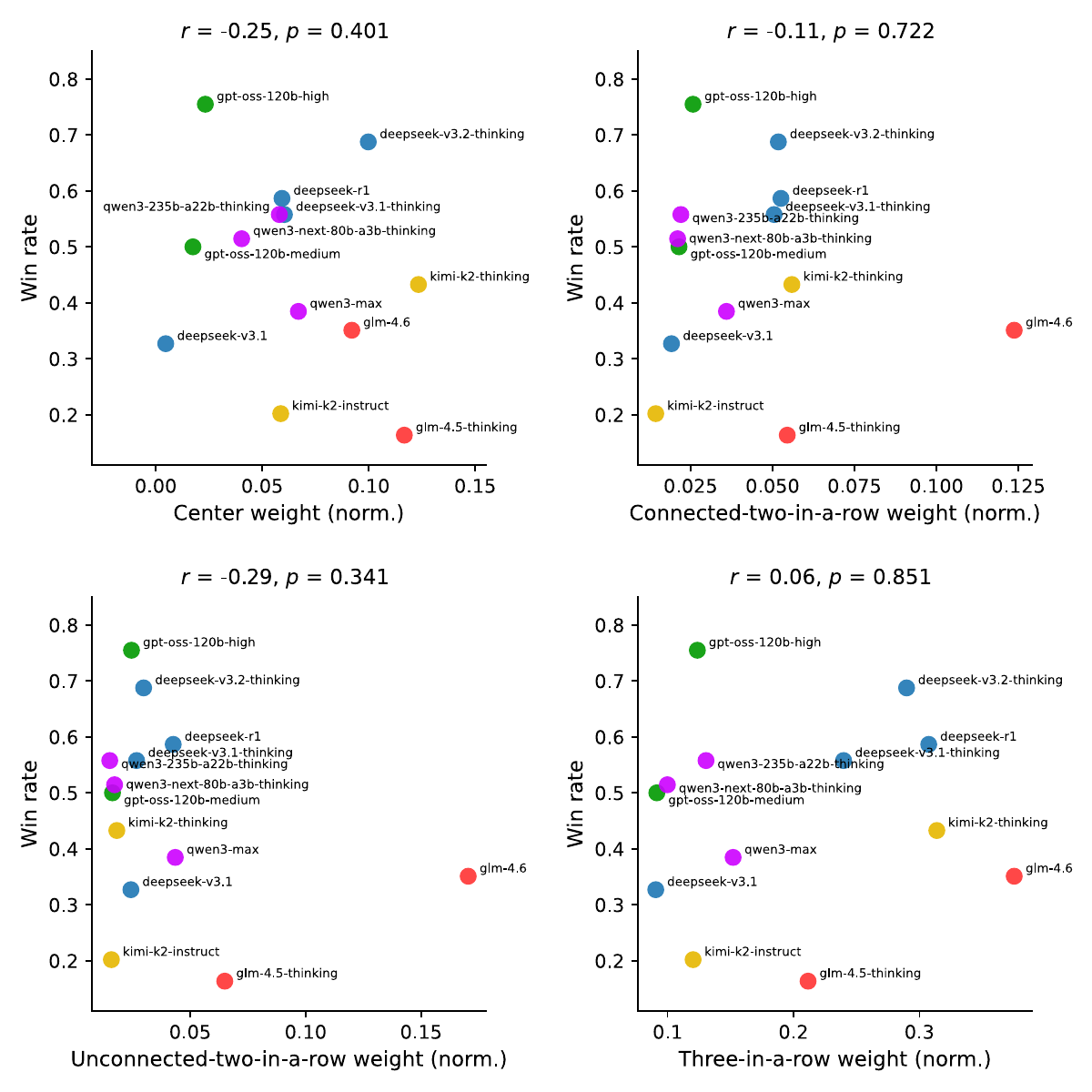}
    \caption{
    \textbf{Normalized feature weights versus win rate.}
    }
    \label{fig:sf3}
\end{figure}

Each panel shows the relationship between one feature weight (normalized by the four-in-a-row weight) and win rate across models. Normalization removes the overall scale of the value function, isolating each LLM's relative preference for a given feature. None of the four normalized weights correlated significantly with win rate (\autoref{fig:sf3}).

\subsection{Robustness to the choice of heuristic features}

The cognitive models in the main text evaluate board states with a particular heuristic value function. One concern is that this feature set might favor the myopic model. If the human-derived features capture only what matters at shallow depth, deep search could appear harmful because the value function is miscalibrated. To test this, we refit all model variants with a generic feature set that makes no assumption about which piece patterns are relevant.

We enumerated all length-4 windows on the 4$\times$9 board (horizontal, vertical and diagonal), and described each window only by its composition, i.e., the number of pieces belonging to the active player and the number belonging to the opponent. Every composition with between one and four pieces in total was assigned its own count feature, yielding 14 composition features. For example, these features include (1 own, 0 opponent), (2 own, 1 opponent) and (0 own, 3 opponent). Together with the center feature, this gave 15 features in total. A composition and its mirror, such as (3, 0) and (0, 3), received separate weights. These weights can express any asymmetry between attack and defense directly, so we fixed the scaling parameter $C$ to 1.

\begin{figure}[h]
    \centering
    \includegraphics[width=\linewidth]{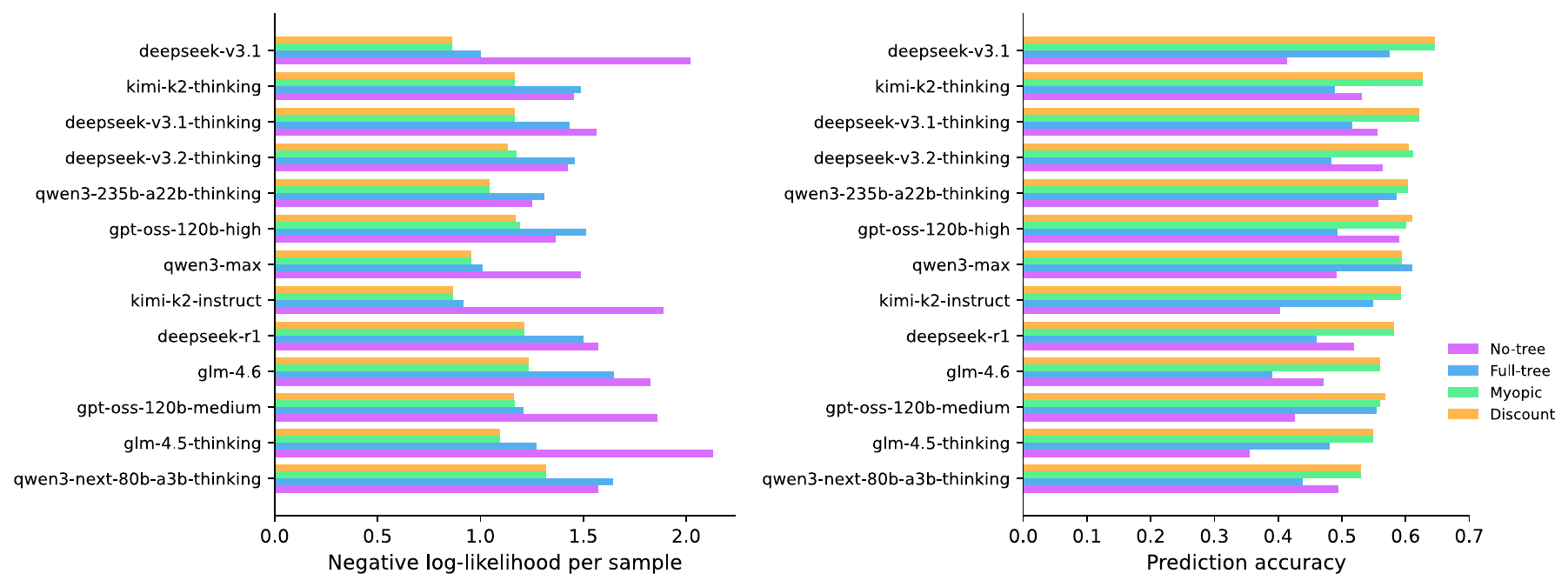}
    \caption{
    \textbf{Model comparison with the window-count feature set.}
    Models are sorted by the accuracy of the myopic model.
    }
    \label{fig:sf4}
\end{figure}

The main-text conclusions hold with the new features (\autoref{fig:sf4}). The myopic model outperformed the no-tree baseline in all 13 LLMs, suggesting that the candidate moves extracted from the reasoning traces remain predictive of the chosen move. The myopic model outperformed the full-tree model in every LLM, and achieved higher prediction accuracy in 12 of the 13 LLMs. In the discount model, the fitted discount factor converged to 0 in 10 of 13 models, which reduces the discount model to the myopic model. For the remaining three LLMs, it stayed small: 0.21 for GPT-OSS-120B (high), 0.23 for DeepSeek-V3.2-Thinking, and 0.44 for GPT-OSS-120B (medium). The information from deeper in the tree thus received little weight. Together, the preference for the myopic model is not an artifact of the human-derived features. It persists when the value function is learned from each LLM's own choices over a general feature set.

\end{document}